\documentclass[journal]{IEEEtran}

\ifCLASSINFOpdf
\else
   \usepackage[dvips]{graphicx}
\fi
\usepackage{url}

\usepackage{graphicx}

\usepackage{xcolor}
\usepackage{algorithm}
\usepackage{algcompatible}
\algnewcommand\INPUT{\item[\textbf{Input:}]}%
\algnewcommand\RETURN{\item[\textbf{Return:}]}%
\usepackage{comment}
\usepackage{kotex}
\usepackage{amssymb}
\usepackage{amsmath}
\usepackage{cleveref}
\usepackage{booktabs}
\usepackage{colortbl}
\usepackage{multirow}
\usepackage{arydshln}
\usepackage{pifont}
\usepackage{microtype}
\usepackage{cite}
\usepackage{balance}
\newcommand{\tabref}[1]{Table~\ref{#1}}
\newcommand{\figref}[1]{Figure~\ref{#1}}
\newcommand{\secref}[1]{Sec.~\ref{#1}}

\begin{document}

% \title{Preparation of Papers for IEEE Signal Processing Letters (5-page limit)}

% \title{Rethinking 3D Occupancy World Models with Cascaded Discrete Representations}
\title{CascadeOcc: Rethinking 3D Occupancy World Models with Cascaded VQ Representations}

\author{Kyumin Hwang$^{*}$, Wonhyeok Choi$^{*}$, Jaeyeul Kim, Jihun Park, Daehee Park$^{\dagger}$, and~Sunghoon~Im$^{\dagger}$,~\IEEEmembership{Member,~IEEE}
\thanks{Manuscript received: January 6, 2026; Revised: March 17, 2026; Accepted: March 30, 2026. This work was supported by Korea Research Institute for defense Technology planning and advancement through Defense Innovation Vanguard Enterprise Project, funded by Defense Acquisition Program Administration (R230206), the Technology Innovation Program (RS-2024-00445759, Development of Navigation Technology Utilizing Visual Information Based on Vision-Language Models for Understanding Dynamic Environments in Non-Learned Spaces) funded by the Ministry of Trade, Industry \& Energy (MOTIE, Korea) and LG AI STAR Talent Development Program for Leading Large-Scale Generative AI Models in the Physical AI Domain (RS-2025-25442149).}
\thanks{K. Hwang, W. Choi, J. Kim, J. Park, D. Park, S. Im with the Daegu Gyeongbuk Institute of Science and Technology (DGIST), Daegu 42988, South Korea. E-mail: \{kyumin, smu06117, jykim94, pjh2857, dhpark, sunghoonim\}@dgist.ac.kr}% <-this % stops a space
\thanks{*: Equal Contribution, $\dagger$: D. Park and S. Im are the Corresponding authors.}}
% \thanks{$\dagger$: S. Im is the Corresponding author.}}
% \thanks{This paragraph of the first footnote will contain the date on which you submitted your paper for review. It will also contain support information, including sponsor and financial support acknowledgment. For example, ``This work was supported in part by the U.S. Department of Commerce under Grant BS123456.'' }
% \thanks{The next few paragraphs should contain the authors' current affiliations, including current address and e-mail. For example, F. A. Author is with the National Institute of Standards and Technology, Boulder, CO 80305 USA (e-mail: author@boulder.nist.gov).}
% \thanks{S. B. Author, Jr., was with Rice University, Houston, TX 77005 USA. He is now with the Department of Physics, Colorado State University, Fort Collins, CO 80523 USA (e-mail: author@lamar.colostate.edu).}}

\markboth{Journal of \LaTeX\ Class Files, Vol. 14, No. 8, August 2015}
{Shell \MakeLowercase{\textit{et al.}}: Bare Demo of IEEEtran.cls for IEEE Journals}
\maketitle

\begin{abstract}

This letter proposes CascadeOcc, a novel occupancy world model that prioritizes intrinsic structural hierarchy over extrinsic auxiliary modalities for autonomous driving.
Occupancy world models---forecasting the future driving environment and planning the driving trajectory---effectively bridge perception and planning, but current approaches often heavily rely on external modalities or large language models, failing to fully exploit the inherent structural potential of occupancy representations themselves.
To enhance representational capacity for complex 3D scenes, we integrate a cascaded Vector Quantized (VQ) mechanism into an autoregressive framework.
Following a coarse-to-fine principle, CascadeOcc progressively refines fine-grained details from global structures through a multi-scale architecture.
Additionally, we incorporate a TimeMixer to capture multi-scale temporal dependencies, establishing a dual-hierarchy mechanism in both space and time.
Experimental results on 4D occupancy forecasting and motion planning benchmarks demonstrate that CascadeOcc achieves superior performance among vision-centric approaches, validating that optimizing inherent representations is a powerful alternative to relying on external foundation models.
% The code will be made available at \url{https://github.com/KyuminHwang/CascadeOcc} upon acceptance.
\end{abstract}


\IEEEpeerreviewmaketitle

\section{Introduction}
\label{sec:introduction}
\IEEEPARstart{T}{he} paradigm of 3D scene understanding in autonomous driving is being reshaped around representations centered on bird's-eye view (BEV)~\cite{yang2023bevformer, huang2021bevdet} and occupancy~\cite{cao2022monoscene, li2023voxformer}, marking a departure from traditional dense depth estimation approaches~\cite{guizilini2022full, wei2023surrounddepth, schmied2023r3d3, xu2021monocular, li2021adv}. 
A major catalyst for this shift has been the emergence of BEV representation based methodologies such as BEVFormer~\cite{li2024bevformer} and LSS~\cite{philion2020lift}, which infer 3D structure from multi-camera images without relying on dense depth maps, along with occupancy prediction methods extended from these representations.
In particular, the introduction of dense occupancy label datasets, as proposed in SurroundOcc~\cite{wei2023surroundocc}, Occ3D~\cite{tian2023occ3d}, and OpenOccupancy~\cite{wang2023openoccupancy}, along with the tri-plane representation introduced in TPVFormer~\cite{huang2023tri} that effectively integrates the strengths of voxel and BEV representations, has played a significant role in accelerating the progress of vision-based occupancy prediction to a level that is comparable to LiDAR-based perception.

Building on this success, recent research has begun to move beyond static 3D scene understanding toward forecasting future driving environments and vehicle trajectories, giving rise to the concept of the occupancy world.
OccWorld~\cite{zheng2024occworld}, an early work in the occupancy world framework, transforms 3D scenes into discrete tokens using a Vector Quantized Variational AutoEncoder (VQVAE)~\cite{van2017neural}, and forecasts future scene occupancy and ego-vehicle trajectory through a spatial-temporal transformer. 
More recently, there has been a growing body of research exploring the integration of world models with large language models (LLM), leverage their rich contextual representations to enable broader scene understanding~\cite{wei2024occllama, xu2025occ}.
% \initial{Alongside this line of work, other approaches have emerged that enhance future prediction by transformers through the use of separate codebooks, which distinguish between movable elements and free-space regions within driving scenes~\cite{xu2025occ, yan2024renderworld}.}
% \suggest{separate codebooks보다 좀 더 가독성 있는 표현으로 쓰는게 좋아 보입니다 예를 들면 movable 여부나 free space여부에 따라서 각각 다른 연산을 사용했다 등}
Alongside this line of work, other approaches have emerged that enhance future prediction by employing distinct encoding mechanisms based on scene semantics, effectively disentangling movable elements from static free-space regions within driving scenes~\cite{xu2025occ, yan2024renderworld}.
Furthermore, methods like FSF-Net~\cite{guo2025fsf} have sought to improve fine-grained 4D occupancy forecasting by explicitly capturing spatial-temporal dynamics through the fusion of coarse BEV scene flow and vector-quantized networks.
However, recent research in the world model has faced key limitations, including the complexity arising from a heavy reliance on auxiliary modalities or external knowledge priors, and structural disjoints caused by artificially partitioning the scene into separate latent spaces, which potentially complicates the holistic modeling of interactions between dynamic agents and their surroundings.
Consequently, prior approaches have not thoroughly investigated the potential of refining structural representations to bridge the occupancy world model and the autoregressive model, nor demonstrated how fully exploiting these inherent capabilities can achieve superior performance without necessitating external knowledge.

We propose a streamlined approach to address these challenges through a novel architecture, CascadeOcc.
Specifically, we design CascadeOcc to seamlessly incorporate the cascade paradigm, or coarse-to-fine strategy, which has penetrated advances in the multi-view stereo and the autoregressive model, into the occupancy world model.
Inspired by the planning-driven philosophy of UniAD~\cite{hu2023planning}, we adopt a progressively cascading design as the core principle of our framework. 
Specifically, we leverage a hierarchical multi-scale VQVAE-v2~\cite{razavi2019generating} to maximize both the spatial expressiveness and reconstruction fidelity of 3D voxel representations.
% Our approach tokenizes input 3D occupancy voxels into multi-scale discrete tokens and reconstructs them hierarchically. 
% At each scale, the output of the previous scale can act as conditional guidance for the next scale, robust structural consistency, while preserving the high-fidelity reconstruction capabilities of the autoregressive model.
At each scale, the output of the previous scale can serve as conditional guidance for the following scale, ensuring robust structural consistency while preserving the high-fidelity reconstruction capabilities of the autoregressive model.
% To anchor these expressive multi-scale voxel tokens in the future forecasting of driving environments, we introduce a cascaded spatial-temporal transformer architecture.
% By applying a progressive sharpening approach that reconstructs the overall driving scene at a coarse scale and then gradually focuses on dynamic fine elements at a finer scale, we achieved a cost-effective cascade occupancy world model that is both accurate and predictive.
This structure enables a progressive refinement process, wherein a coarse reconstruction of the global scene is followed by an increasingly detailed focus on dynamic and fine-grained elements. 
% This design leads to a predictive and cost-efficient cascade occupancy world model.
% Furthermore, we proposed TimeMixer, a dedicated module within the transformer that extends the coarse-to-fine design, which was limited to the sptial dimension, to the temporal domain for predicting the planning scenario of autonomous driving, and effectively fuses long-range and short-range dependencies.
Furthermore, we propose TimeMixer to establish a dual-hierarchy mechanism across both spatial and temporal dimensions. By extending the coarse-to-fine design to the temporal domain, TimeMixer effectively fuses long-range and short-range dependencies for precise planning scenario prediction.

% Moreover, we extened the coarse-to-fine paradigm, traditionally confined to the spatial domain, into the temporal dimensions.
% For this, we propose TimeMixer, facilitates the fusion of short-term and long-term dependencies, thereby enhancing planning-oriented temporal reasoning for autonomous driving.

\begin{figure*}[!t]
\centering
\includegraphics[width=0.87\linewidth]{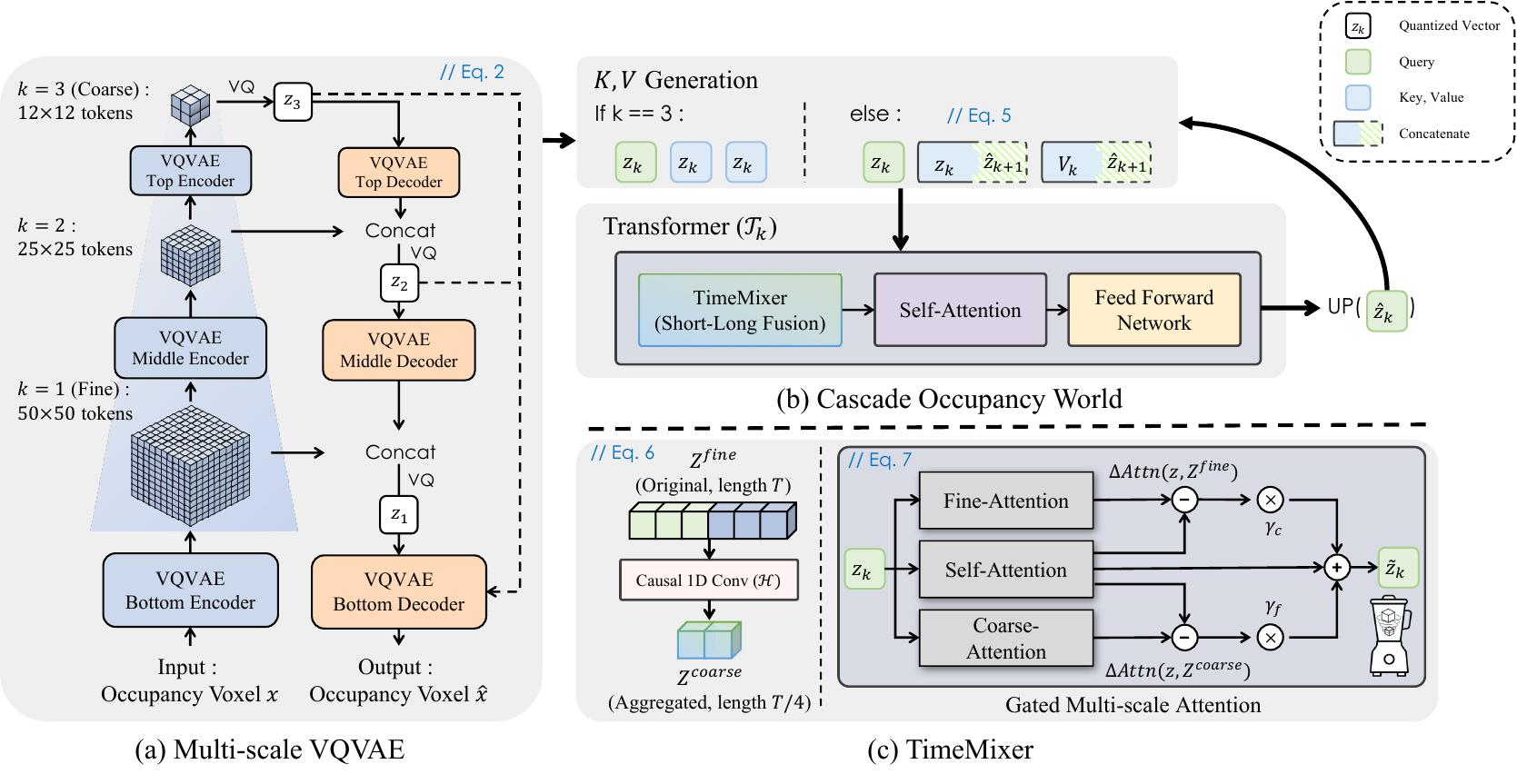} % Reduce the figure size so that it is slightly narrower than the column. Don't use precise values for figure width.This setup will avoid overfull boxes.
\caption{\textbf{Structure of CascadeOcc.} Given a sequence of 3D occupancy inputs, the \textbf{Multi-scale VQVAE} (a) first encodes the scene into hierarchical discrete tokens. The \textbf{Cascade Occupancy World} (b) then progressively forecasts future states from coarse to fine levels. To capture complex temporal dynamics, the \textbf{TimeMixer} (c) adaptively aligns short- and long-term contexts using gated attention, guiding the model to generate high-fidelity future occupancy predictions.}
\label{fig:pipeline}
\vspace{-5pt}
\end{figure*}

% To demonstrate the effectiveness of CascadeOcc, Occ3D~\cite{tian2023occ3d}, nuScenes~\cite{caesar2020nuscenes}
To demonstrate the effectiveness of CascadeOcc, we evaluate its performance on 4D occupancy forecasting and ego-motion planning using the Occ3D~\cite{tian2023occ3d} and nuScenes~\cite{caesar2020nuscenes} datasets.
In 3D occupancy reconstruction, CascadeOcc achieves notable improvements over OccWorld, with an increase of 2.24\% in IoU and 4.6\% in mIoU.
These results indicate that the proposed method effectively preserves rich representations within 3D occupancy voxels.
% Furthermore, in 4D occupancy forecasting, CascadeOcc achieves \textcolor{red}{3.65}\% and \textcolor{red}{3.2}\% improvements in IoU and mIoU, respectively, over OccWorld, while maintaining a similar computational cost of \textcolor{red}{XX.XX} FPS, comparable to the 18 FPS of OccWorld. It also shows significant gains in planning-relevent metrics achieving \textcolor{red}{XX.XX}\% and \textcolor{red}{XX.XX}\% improvements in L2 error and collision rate, respectively. Theses results highlight the reliable planning ability of the proposed occupancy world model.
% Furthermore, in 4D occupancy forecasting, CascadeOcc achieves \textcolor{red}{3.65}\% and \textcolor{red}{3.2}\% improvements in IoU and mIoU, respectively, over OccWorld, while even achieving a lower collision rate, thereby ensuring safer motion planning.
Furthermore, in 4D occupancy forecasting, CascadeOcc achieves 3.65\% and 3.2\% improvements in IoU and mIoU, respectively, over OccWorld. 
Moreover, our model demonstrates enhanced safety by achieving a lower collision rate, highlighting the reliable planning capability of the proposed occupancy world model. 
Our contributions are threefold as follows:
\begin{itemize}
\item To the best of our knowledge, we are the first to introduce a novel Occupancy World Model, CascadeOcc, which seamlessly integrates a cascaded VQ representation into an autoregressive framework.
\item We propose TimeMixer, which extends the coarse-to-fine paradigm to the temporal domain to establish a dual-hierarchy mechanism, effectively fusing long- and short-range dependencies for accurate forecasting and planning.
\item Extensive experiments demonstrate that CascadeOcc achieves superior performance in both forecasting and planning without reliance on auxiliary knowledge, validating that optimized inherent representations are sufficient to ensure enhanced user safety.
% \item Extensive experiments demonstrate that CascadeOcc achieves state-of-the-art performance, not only in forecasting accuracy but also in planning tasks, thereby ensuring enhanced user safety.
\end{itemize}
\section{Method}
\label{sec:method}
The overall pipeline of the proposed \textit{CascadeOcc} is illustrated in \figref{fig:pipeline}.
The primary objective of the proposed method is to design an occupancy world model that maximizes the expressive capacity of the autoregressive model for 4D occupancy forecasting and motion planning, without relying on additional information or modalities. 
Consistent with the OccWorld framework, we adopt a two-stage training paradigm, including occupancy reconstruction and forecasting phases, and our discussion primarily concentrates on the distinctive structural innovations.
Furthermore, while our framework jointly estimates both occupancy tokens and ego-pose, we omit the explicit mathematical formulation for pose estimation for the sake of brevity, as these operations remain identical to the baseline implementation.
% To achieve this, we systematically decompose the proposed CascadeOcc into three components as shown in ~\figref{fig:pipeline}.
% In Section \ref{sec:vqvae-v2}, we present the formulation of the multi-scale VQVAE. 
% In Section \ref{sec:cascade-tf}, we describe how the multi-scale discrete tokens are seamlessly adopted in a coarse-to-fine manner within a transformer architecture for future prediction.
% In Section \ref{sec:timemixer}, we introduce the proposed TimeMixer, which extends the cascade paradigm along the temporal axis to effectively capture the progression of time.

\subsection{Multi-scale Scene Tokenizer Formulation}
\label{sec:vqvae-v2}
To enhance the representational capacity for complex driving scenes, we incorporate a hierarchical multi-scale VQVAE-v2~\cite{razavi2019generating} into our scene tokenizer. 
Given a 3D occupancy voxel $x \in \mathbb{R}^{H \times W \times D}$, we first transform it into a feature embedding $x_{emb} \in \mathbb{R}^{H \times W \times (D \cdot C)}$ via learnable $C$ class embeddings.
% Here, $H, W, \text{and } D$ denote the spatial resolution of the voxel space, and $C$ represents the dimension of the learnable class embedding assigned to each semantic category.
The encoder operates progressively from the highest resolution ($k=1$) to the lowest ($k=3$).
Encoder features $e_k$ at each level are computed as:
% To preserve environmental context from multiple perspectives, the encoding process operates in a fine-to-coarse manner, progressively downsampling the spatial resolution to aggregate global semantics. 
% Specifically, the encoding starts from the bottom scale (highest resolution, $k=0$) and proceeds to the top scale (lowest resolution, $k=2$). 
% The encoder feature $e_k$ at scale $k \in \{0, 1, 2\}$ is extracted sequentially from the feature of the previous scale:
\begin{equation}
e_k = \mathcal{E}_k(e_{k-1}), \quad \text{where } e_{0} = x_{emb}.
\label{eq:1}
\end{equation}
At each scale $k$, the spatial resolution is downsampled by a factor of $2^{k-1}$ to aggregate global semantics.
% Here, $\mathcal{E}_k$ denotes the encoder block at scale $k$.
% As the scale level $k$ increases, the spatial resolution decreases by a factor of $2^k$. 

% Conversely, the subsequent quantization proceeds in a coarse-to-fine manner.
% The discrete token $z_k$ is obtained by conditioning the current encoder feature $e_k$ on the upsampled reconstruction $\hat{h}_{k+1}$ from the coarser scale.
Conversely, the subsequent quantization proceeds from coarser to finer scales to preserve structural details.
For the top scale, the discrete token $z_3$ is derived directly from $e_3$.
For finer scales tokens $\{z_2, z_1\}$ are obtained by conditioning $\{e_3, e_2\}$ on the upsampled reconstruction $\hat{h}_{k+1}$ from the coarser scale:
% Specifically, for the top scale ($k=2$), the discrete token $z_k$ is derived directly from $e_k$. For finer scales ($k < 2$), $z_k$ is obtained by conditioning $e_k$ on the upsampled reconstruction $\hat{h}_{k+1}$ from the coarser scale:
\begin{equation}
z_k = \begin{cases} \mathcal{Q}_k(e_k) & \text{if } k=3 \\
\mathcal{Q}_k(\text{Concat}(e_k, \hat{h}_{k+1})) & \text{else} \end{cases}, \quad \hat{h}_k = \mathcal{D}_k(z_k),
\label{eq:2}
\end{equation}
where $\mathcal{Q}_k$ and $\mathcal{D}_k$ denote the quantizer and decoder blocks at scale $k$.
Inspired by the Feature Pyramid Network (FPN)~\cite{lin2017feature}, we fuse all multi-scale quantized tokens $\{z_1, z_2, z_3\}$ into the bottom decoder to reconstruct the 3D occupancy scene as illustrated in \figref{fig:pipeline}-(a). 
This hierarchical formulation is meticulously designed to capture both global context and fine-grained details, ensuring robust modeling of dynamic autonomous driving environments.
% Going beyond a simple application of VQVAE-v2, this formulation is meticulously designed to maximize the representational capacity of occupancy, allowing for robust modeling of dynamic autonomous driving environments.

\subsection{Cascade Occupancy World Model}
\label{sec:cascade-tf}
% Unlike UniAD~\cite{hu2023planning}, which treats prediction and planning as separate tasks, our Cascade Occupancy World Model adheres to the OccWorld framework, jointly performing 4D occupancy forecasting and ego-vehicle planning.
Following the OccWorld, our model jointly performs 4D occupancy forecasting and ego-vehicle planning.
% However, we diverge from OccWorld’s approach, which relies on a subsequent 2D U-Net for multi-resolution features.
% Instead, inspired by coarse-to-fine generation paradigms like VQVAE-v2, we integrate a cascading strategy directly into the forecasting transformers.
However, instead of using a subsequent 2D U-Net for multi-resolution features, we integrate a cascading strategy directly into the forecasting transformers, inspired by coarse-to-fine generation paradigms~\cite{razavi2019generating}.

Let $Z_k = \{z_k^t\}_{t=1}^T$ and $P_k = \{p_k^t\}_{t=1}^T$ denote the observed sequence of discrete occupancy tokens and ego-poses at scale $k \in \{1, 2, 3\}$ of length $T$. 
% The forecasting process operates sequentially from the coarsest scale (top, $k=2$) to the finest (bottom, $k=0$). 
The forecasting proceeds sequentially from the coarsest scale ($k=3$) to finest ($k=1$).

% \revision{
To predict the future state at $T+1$, we employ a scale-specific transformer $\mathcal{T}_k$ equipped with a guidance-aware attention mechanism.
Query $Q_k$ is derived from the current scale's temporal context $Z_k$, while key $K_k$ and value $V_k$ are projected from an augmented context sequence $C_k$ to incorporate hierarchical guidance:
% Inside the transformer $\mathcal{T}_k$, we construct the Query ($Q_k$), Key ($K_k$), and Value ($V_k$) matrices to maintain temporal consistency while incorporating hierarchical guidance. 
% Specifically, $Q_k$ is derived solely from its own scale-specific temporal context $Z_k$, whereas $K_k$ and $V_k$ are projected from an augmented context sequence $C_k$:
\begin{gather}
Q_k = Z_k W^Q_k, \quad K_k, V_k = C_k W^{K,V}_k,\\
% \label{eq:4}
% \end{equation}
% \begin{equation}
(\hat{z}^{T+1}_k, \hat{p}^{T+1}_k) = \mathcal{T}_k \left( \text{Attn}(Q_k, K_k, V_k) \right),
\label{eq:34}
\end{gather}
where $W^Q_k$ and $W^{K,V}_k$ are learnable projection matrices for scale $k$, and the augmented context sequence $C_k = \{c^t_k\}_{t=1}^T$ is constructed by concatenating the temporal context of the current scale with the upsampled prediction from the coarser scale:
% This asymmetric attention design ensures that the model queries the specific dynamics of the current resolution ($Q_k$) by referencing both its own temporal context and the structural guidance from the coarser level ($K_k, V_k$).
\begin{equation}
c_k^t = \begin{cases} z_k^t & \text{if } k=3 \\
\text{Concat}(z_k^t, \mathcal{U}(\hat{z}_{k+1}^{T+1})) & \text{else} \end{cases},
\label{eq:5}
\end{equation}
where $\hat{z}^{T+1}_{k+1}$ is the latent token previously forecasted by the coarser scale transformer $\mathcal{T}_{k+1}$, and $\mathcal{U}$ denotes the upsampling operation. 
% This mechanism ensures that the model first establishes a global scene structure (e.g., static background) and progressively refines dynamic, fine-grained details, resulting in more robust forecasting and reliable planning in complex environments.
Instead of forecasting the entire scene at once, our approach first establishes the global context (e.g., background) and progressively modifies fine-grained details, resulting in more robust forecasting and reliable driving plans. 
% This hierarchical refinement allows for robust forecasting of dynamic and challenging future scenes, leading to more reliable autonomous driving plans.
% }

% \subsection{TimeMixer: Temporal-Hierarchy from Long- to Short-Range}
\subsection{\textls[-10]{TimeMixer: Temporal-Hierarchy from Long- to Short-Range}}
\label{sec:timemixer}
While this cascading strategy establishes a spatial hierarchy, we propose TimeMixer to extend this paradigm to the temporal domain.
% we propose TimeMixer to extend this paradigm to the temporal domain, achieving a dual-hierarchy mechanism that captures both spatial and temporal multi-scale dependencies.
% \revision{Specifically, this module is employed in the scale-specific transformer operation defined in ~\eqref{eq:7}, with its detailed architecture illustrated in ~\figref{fig:pipeline}(c).}
% -> 후에 intial 지우고 eqref 체크할것}
Inspired by the coarse-to-fine hypothesis in CasMVSNet~\cite{gu2020cascade}, TimeMixer introduces a temporal pyramid within the scale-specific transformer to balance long- and short-range dynamics.
% within the transformer.
We encode the input token and pose sequences $Z = \{z^{t}\}_{t=1}^T$ and $P = \{p^{t}\}_{t=1}^T$ using a causal 1D convolutional block $\mathcal{H}_{\text{causal}}$~\cite{van2016wavenet}, consisting of two stacked convolutions (kernel size 2, stride 2).
This operation extracts a coarse representation by reducing the temporal resolution by a factor of four:
\begin{equation}
Z^{coarse} = \mathcal{H}(Z), \quad P^{coarse} = \mathcal{H}(P).
\label{eq:6}
\end{equation}
The coarse features enable effective long-range encoding by filtering high-frequency noise, while the original sequences ($Z^{fine} \equiv Z, P^{fine} \equiv P$) retain fine-grained resolution for short-range interactions.

% We encode the temporal axis of the discrete token sequence $Z$ and pose sequence $P$ using a causal 1D convolutional block~\cite{van2016wavenet} $\mathcal{H}_{\text{causal}}$.
% This block consists of two stacked 1D convolutions (kernel size 2, stride 2) with activation functions, aggregating coarse representations:
% \begin{equation}
%     Z^{coarse} = \mathcal{H}_{\text{causal}}(Z),\; P^{coarse} = \mathcal{H}_{\text{causal}}(Z)
% \label{eq:6}
% \end{equation}
% This design reduces the temporal resolution by a factor of four, thereby limiting indiscriminate exploration over the entire past scene and enabling effective long-range encoding that facilitates the discovery of meaningful long-term dependencies relevant to the current scene dynamics. 
% Conversely, the original sequences ($Z^{fine} \equiv Z$ and $P^{fine} \equiv P$) retain fine-grained temporal resolution, serving as the source for short-range interactions.

To effectively integrate these dual-scale features, we design a gated attention module. 
The final temporal context $\tilde{z}$ is computed by fusing the self-attention output with cross-scale residual gains:
% guided attention from both coarse and fine scales:
\begin{equation}
\small
    \tilde{z} = \text{Attn}_{self}(z) + \gamma_c \Delta \text{Attn}(z, Z^{coarse}) + \gamma_f \Delta \text{Attn}(z, Z^{fine}),
\label{eq:7}
\end{equation}
where $z$ is the query at the current time step, $\Delta \text{Attn}(Q, K) = \text{Attn}(Q, K) - \text{Attn}_{self}(Q)$ represents the residual information gain from cross-scale attention, and $\gamma_c, \gamma_f$ are learnable gating parameters. 
% This formulation allows the model to dynamically balance global temporal context (coarse) and local motion details (fine) for accurate trajectory planning and forecasting.
This formulation allows CascadeOcc to dynamically weigh global temporal context against local motion details, ensuring accurate trajectory planning and robust forecasting in complex environments.

\begin{figure}[!t]
    \centering
    \includegraphics[width=1.0\linewidth]{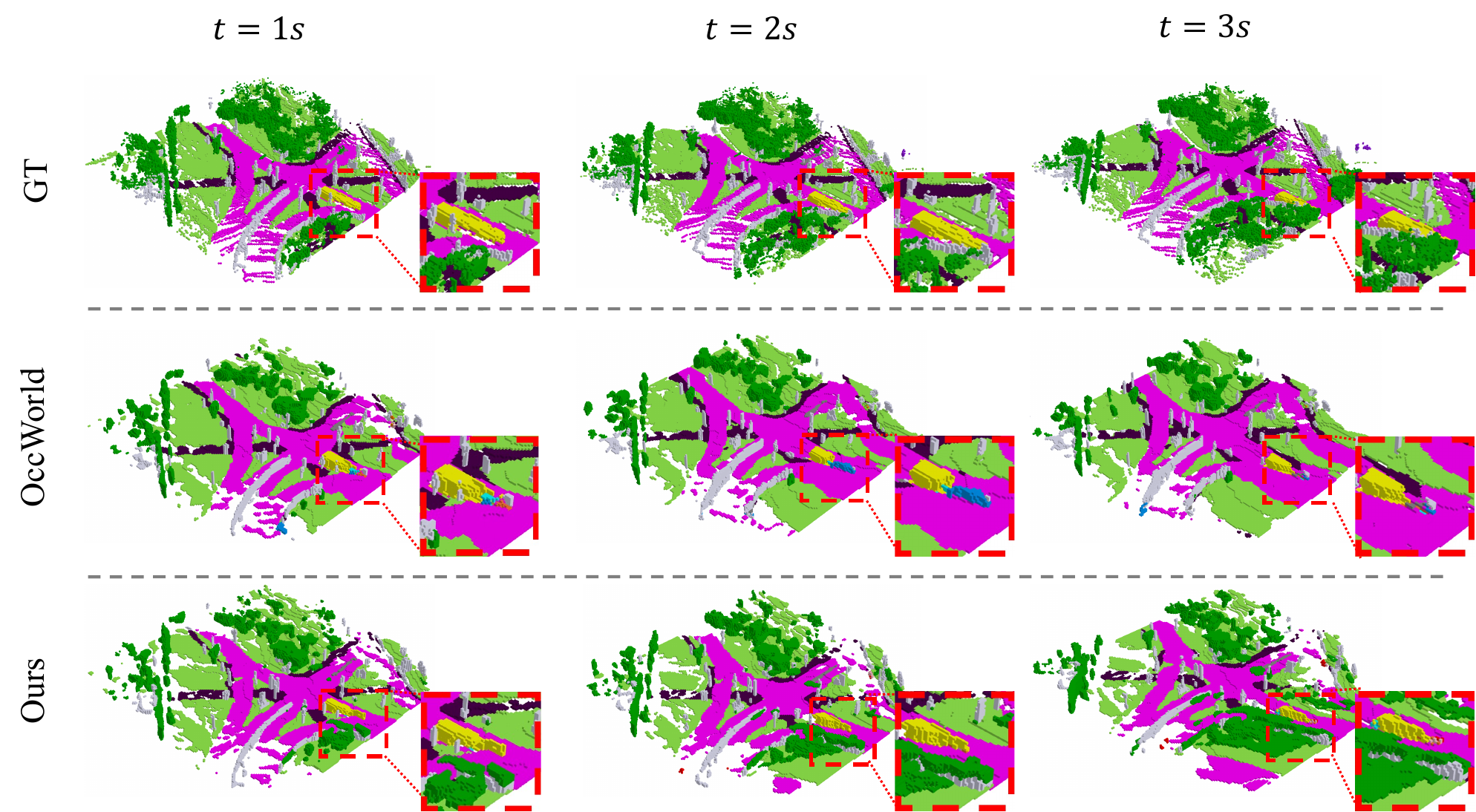}
    \caption{Qualitative results of the forecasting and planning with CascadeOcc. Best viewed \textbf{ZOOMED-IN}.}
    \label{fig:qualitative}
    % \vspace{-10pt}
\end{figure}

\begin{table*}[t]
\centering
\caption{3D occupancy reconstruction performance on the Occ3D-nuScenes validation dataset.}
\label{tab:reconstruction}
\setlength{\aboverulesep}{0.2pt}
\setlength{\belowrulesep}{1.0pt}
\resizebox{\textwidth}{!}{%
\setlength{\tabcolsep}{3pt} % 간격을 조금 더 넓혀서 가독성 확보
\renewcommand{\arraystretch}{1.2}
% 대각선 텍스트를 위한 매크로 정의 (45도)
\newcommand{\diagcolumn}[1]{\multicolumn{1}{l}{\rotatebox{55}{\makebox[0pt][l]{\scriptsize#1}}}}

\begin{tabular}{l cc ccccccccccccccccc}
\toprule
\addlinespace[21pt] % 대각선 텍스트가 위쪽 라인과 겹치지 않게 공간 확보
\textbf{Methods} & \textbf{IoU $\uparrow$} & \textbf{mIoU $\uparrow$} &
\diagcolumn{Others} & \diagcolumn{barrier} & \diagcolumn{bicycle} & \diagcolumn{bus} & \diagcolumn{car} & \diagcolumn{cons. veh} & \diagcolumn{motorcycle} & \diagcolumn{pedestrian} & \diagcolumn{traffic cone} & \diagcolumn{trailer} & \diagcolumn{truck} & \diagcolumn{dri. sur} & \diagcolumn{other flat} & \diagcolumn{sidewalk} & \diagcolumn{terrain} & \diagcolumn{manmade} & \diagcolumn{vegetation} \\
\addlinespace[-1pt] % 대각선 텍스트가 위쪽 라인과 겹치지 않게 공간 확보
\midrule
\addlinespace[-1pt] % 대각선 텍스트가 위쪽 라인과 겹치지 않게 공간 확보
OccWorld~\cite{zheng2024occworld} & 61.88 & 64.74 & 46.34 & 71.84 & 69.96 & 67.59 & 69.01 & 45.14 & 73.50 & 74.77 & 68.57 & 54.65 & 65.27 & 82.74 & 78.18 & 69.81 & 66.53 & 52.95 & 43.77 \\
\addlinespace[-2pt] % 대각선 텍스트가 위쪽 라인과 겹치지 않게 공간 확보
\midrule
\addlinespace[-1pt] % 대각선 텍스트가 위쪽 라인과 겹치지 않게 공간 확보
\textbf{CascadeOcc (Ours)} & \textbf{64.12} & \textbf{69.34} & \textbf{54.82} & \textbf{77.90} & \textbf{76.41} & \textbf{72.27} & \textbf{72.96} & \textbf{51.07} & \textbf{79.21} & \textbf{79.27} & \textbf{75.34} & \textbf{61.25} & \textbf{68.48} & \textbf{84.16} & \textbf{82.00} & \textbf{72.37} & \textbf{69.64} & \textbf{55.35} & \textbf{46.32} \\
\addlinespace[-1pt] % 대각선 텍스트가 위쪽 라인과 겹치지 않게 공간 확보
\bottomrule
\end{tabular}%
}
\vspace{-10pt}
\end{table*}

% \begin{table*}[t]
% \centering
% \caption{3D occupancy reconstruction performance on the Occ3D-nuScenes validation dataset.}
% \label{tab:reconstruction}
% \setlength{\aboverulesep}{0.2pt}
% \setlength{\belowrulesep}{1.0pt}
% \resizebox{\textwidth}{!}{%
% \setlength{\tabcolsep}{1.5pt}
% \renewcommand{\arraystretch}{1.2}
% \begin{tabular}{l cc ccccccccccccccccc}
% \toprule
% \textbf{Methods} & \textbf{IoU $\uparrow$} & \textbf{mIoU $\uparrow$} &
% \rotatebox{90}{Others} & \rotatebox{90}{barrier} & \rotatebox{90}{bicycle} & \rotatebox{90}{bus} & \rotatebox{90}{car} & \rotatebox{90}{cons. veh} & \rotatebox{90}{motorcycle} & \rotatebox{90}{pedestrian} & \rotatebox{90}{traffic cone} & \rotatebox{90}{trailer} & \rotatebox{90}{truck} & \rotatebox{90}{dri. sur} & \rotatebox{90}{other flat} & \rotatebox{90}{sidewalk} & \rotatebox{90}{terrain} & \rotatebox{90}{manmade} & \rotatebox{90}{vegetation} \\
% \midrule
% OccWorld~\cite{zheng2024occworld} & 61.88 & 64.74 & 46.34 & 71.84 & 69.96 & 67.59 & 69.01 & 45.14 & 73.50 & 74.77 & 68.57                                                       & 54.65 & 65.27 & 82.74 & 78.18 & 69.81 & 66.53 & 52.95 & 43.77 \\
% \midrule
% \textbf{CascadeOcc (Ours)} & \textbf{64.12} & \textbf{69.34} & \textbf{54.82} & \textbf{77.90}  & \textbf{76.41} & \textbf{72.27} & \textbf{72.96} & \textbf{51.07} & \textbf{79.21} & \textbf{79.27} & \textbf{75.34} & \textbf{61.25} & \textbf{68.48} & \textbf{84.16} & \textbf{82.00} & \textbf{72.37} & \textbf{69.64} & \textbf{55.35} & \textbf{46.32} \\
% \bottomrule
% \end{tabular}%
% }
% \end{table*}

\begin{table*}[t]
\centering
\caption{4D occupancy forecasting (mIoU \& IoU) and motion-planning (L2 \& collision rate) performance on the Occ3D-nuScenes dataset. Methods utilizing Large Language Models (LLMs) are marked in gray text. (\textbf{O}: Occupancy, \textbf{M}: Map, \textbf{B}: 3D BBox)}
\label{tab:forecast_planning_performance}
% \resizebox{\linewidth}{!}{% \textwidth -> \linewidth
\setlength{\aboverulesep}{0pt}
\setlength{\belowrulesep}{0pt}
\resizebox{0.99\linewidth}{!}{
\setlength{\tabcolsep}{3pt}
\begin{tabular}{l|c|cccc|cccc|cccc|cccc|cc}
\toprule
% HEADING
\multirow{2}{*}{Method} & \multirow{2}{*}{Input} & \multicolumn{4}{c|}{mIoU (\%) $\uparrow$} & \multicolumn{4}{c}{IoU (\%) $\uparrow$} 
& \multicolumn{4}{c|}{L2 ($m$) $\downarrow$} & \multicolumn{4}{c}{Collision Rate (\%) $\downarrow$}
\\

& & 1s & 2s & 3s & \cellcolor{gray!20}Avg. & 1s & 2s & 3s & \cellcolor{gray!20}Avg.
& 1s & 2s & 3s & \cellcolor{gray!20}Avg. & 1s & 2s & 3s & \cellcolor{gray!20}Avg.
& FPS & Memory
\\

\midrule

OccWorld~\cite{zheng2024occworld}   & \textbf{O}         & 25.78 & 15.14 & 10.51 & \cellcolor{gray!20} 17.14 & 34.63 & 25.07 & 20.18 & \cellcolor{gray!20} 26.63
& \textbf{0.43} & \textbf{1.08} & \textbf{1.99} & \cellcolor{gray!20} \textbf{1.17} & \textbf{0.07} & 0.38 & 1.35 & \cellcolor{gray!20} 0.60
& 10.70 & 15,714
\\

OccNet~\cite{tong2023scene}   & \textbf{O}\&\textbf{M}\&\textbf{B} & - & - & - & - & - & - & - & - 
& 1.29 & 2.31 & 2.98 & \cellcolor{gray!20} 2.25 & 0.20 & 0.56 & \textbf{1.30} & \cellcolor{gray!20} 0.69 
& - & -
\\
\hdashline \noalign{\vskip 2pt}

\color{gray}OccLLaMA~\cite{wei2024occllama}   & \color{gray}\textbf{O}\&\textbf{LLM} & \color{gray}25.05 & \color{gray}19.49 & \color{gray}15.26 & \color{gray}\cellcolor{gray!20} 19.93 & \color{gray}34.56 & \color{gray}28.53 & \color{gray}24.41 & \color{gray}\cellcolor{gray!20} 29.17
& \color{gray} 0.37 & \color{gray} 1.02 & \color{gray} 2.03 & \color{gray}\cellcolor{gray!20} 1.14 & \color{gray} 0.04 & \color{gray} 0.24 & \color{gray} 1.20 & \color{gray}\cellcolor{gray!20} 0.49
& - & -
\\

\color{gray}OccLLM~\cite{xu2025occ}     & \color{gray}\textbf{O}\&\textbf{LLM} & \color{gray}24.02 & \color{gray}21.65 & \color{gray}17.29 & \color{gray}\cellcolor{gray!20} 20.99 & \color{gray}36.65 & \color{gray}32.14 & \color{gray}28.77 & \color{gray}\cellcolor{gray!20} 32.52
& \color{gray} 0.12 & \color{gray} 0.24 & \color{gray} 0.49 & \color{gray}\cellcolor{gray!20} 0.28 & \color{gray} - & \color{gray} - & \color{gray} - & \color{gray}\cellcolor{gray!20} - 
& - & -
\\

\midrule

\textbf{CascadeOcc (Ours)} & \textbf{O} & \textbf{31.17} & \textbf{17.91} & \textbf{11.94} & \cellcolor{gray!20}\textbf{20.34} & \textbf{39.72} & \textbf{28.60} & \textbf{22.50} & \cellcolor{gray!20}\textbf{30.28} 
& \textbf{0.43} & 1.12 & 2.11 & \cellcolor{gray!20} 1.22 & 0.12 & \textbf{0.31} & 1.35 & \cellcolor{gray!20} \textbf{0.59}
& 6.00 & 13,784
\\

\bottomrule
\end{tabular}%
}
\vspace{-5pt}
\end{table*}

\begin{table}[t]
    \caption{Ablation studies of CascadeOcc (\textbf{Best}, \underline{Second-best})}
    \label{tab:ablation}
    \setlength{\aboverulesep}{0.2pt}
    \setlength{\belowrulesep}{1.0pt}
    \centering
    \resizebox{0.80\linewidth}{!}{
    \begin{tabular}{cc|cc|cc}
        \toprule
        \multicolumn{2}{c|}{Components} & \multicolumn{2}{c|}{Forecasting} & \multicolumn{2}{c}{Planning} \\
        \cmidrule(lr){1-2} \cmidrule(lr){3-4} \cmidrule(lr){5-6}
        Cascade & TimeMixer & mIoU & IoU & L2 & Col. Rate \\
        % Cascade & TimeMixer & mIoU (\%) $\uparrow$ & IoU (\%) $\uparrow$ & L2 ($m$) $\downarrow$ & Coll. Rate (\%) $\downarrow$ \\
        \midrule
        \ding{55} & \ding{55} & 26.63 & 17.14 & \textbf{1.17} & 0.60 \\
        \ding{51} & \ding{55} & \underline{28.99} & 18.92 & 1.61 & \textbf{0.58} \\ 
        \ding{55} & \ding{51} & 28.83 & \underline{19.37} & 1.54 & 1.14 \\ 
        \ding{51} & \ding{51} & \textbf{30.28} & \textbf{20.34} & \underline{1.22} & \underline{0.59} \\
        \bottomrule
    \end{tabular}}
\vspace{-10pt}
\end{table}

\section{Experiments}
\subsection{Implementation Details}
% Long Version
% For 3D occupancy reconstruction and 4D occupancy forecasting, we utilize the Occ3D-nuScenes dataset, which is constructed using the label generation pipeline proposed in Occ3D~\cite{tian2023occ3d}.
% In addition, to evaluate the performance of future trajectory planning for ego-vehicles, we measure the L2 error and collision rate using the nuScenes dataset~\cite{caesar2020nuscenes}.
% For a fair comparison with previous works, all training and evaluation protocols strictly follow the implementation details specified in OccWorld.
% Also, the loss functions used to train both the Multi-scale Scene Tokenizer and the Cascaded Occupancy World model follow the design introduced in OccWorld. 
% To mitigate the issue of error accumulation across scales\text{-}commonly referred to as the exposure bias problem~\cite{lee2022autoregressive, ranzato2015sequence}, we adopt a soft-labeling strategy~\cite{lee2022autoregressive} during training. 
% In the CascadeOcc, the Spatial Temporal Transformer is composed of two attention layers each for the pose and token pathways.
% Specifically, the causal 1D convolutional blocks in TimeMixer employ ReLU activations, while the gating parameters in the gated multi-scale attention module are normalized using the Sigmoid function.
% All experiments were conducted using four NVIDIA A6000 GPUs.

% Short Version
We evaluate 3D reconstruction and 4D forecasting on the Occ3D-nuScenes dataset~\cite{tian2023occ3d} using mIoU and IoU, while trajectory planning is assessed on the nuScenes dataset~\cite{caesar2020nuscenes} via L2 error and collision rate. 
For a fair comparison with previous works, all training loss functions and evaluation protocols strictly follow the implementation details specified in OccWorld.
To mitigate the issue of error accumulation across scales\text{-}commonly referred to as the exposure bias problem~\cite{lee2022autoregressive, ranzato2015sequence}, we adopt a soft-labeling~\cite{lee2022autoregressive} during training.
The Spatial Temporal Transformer uses two attention layers per pathway. 
TimeMixer employs ReLU activations, and gating parameters use the Sigmoid function. 
Experiments were conducted on four NVIDIA A6000 GPUs.

\subsection{Performance Evaluation}
\subsubsection{3D Occupancy Reconstruction}
To demonstrate the effectiveness of the proposed approach in complex driving scenarios, we conducted experiments on the Occ3D-nuScenes validation dataset, reporting IoU and mIoU over 17 semantic classes for 3D occupancy reconstruction.
% As shown in Table~\ref{tab:reconstruction}, the proposed method achieves improvements of 2.24\% in IoU and 4.6\% in mIoU compared to OccWorld. These results indicate that the proposed multi-scale scene tokenizer effectively minimizes information loss in the representation of three-dimensional driving environments.
% Notably, the proposed method yields substantial improvements for rare classes, achieving gains of 6.45\% for bicycles, 5.71\% for motorcycles, and 6.77\% for traffic cones. 
% It also demonstrates significant performance improvements for dominant classes such as drivable area and vegetation, with gains of 1.42\% and 2.55\%, respectively.
As shown in~\tabref{tab:reconstruction}, our method outperforms OccWorld with gains of 2.24\% in IoU and 4.6\% in mIoU, effectively minimizing information loss. 
Notably, we achieve substantial improvements in rare classes (e.g., traffic cones +6.77\%, bicycles +6.45\%) as well as dominant classes (e.g., vegetation +2.55\%), demonstrating enhanced representational capacity across diverse semantic categories.
These results suggest that our approach enhances the representational capacity for both dynamic and sparse classes while improving overall understanding of the driving environment.

\subsubsection{4D Occupancy Forecasting}
% To demonstrate the superior forecasting capabilities of our proposed method within dynamic autonomous driving environments, we evaluated our model on the Occ3D-nuScenes dataset. 
% The forecasting task involves predicting the driving environment for the future 3 seconds, conditioned on a historical sequence of the past 2 seconds.
% As reported in \tabref{tab:forecast_planning_performance}, our method achieves significant performance gains over OccWorld, which relies on a single-scale VQVAE. 
% Specifically, we observe improvements of 3.65\% in IoU and 3.2\% in mIoU, respectively. 
% As depicted in ~\figref{fig:qualitative}, these quantitative gains translate into visually superior predictions, where our model accurately captures the evolution of dynamic objects and fine-grained details of static structures.
% Furthermore, the consistent improvements across all future time steps (from 1s to 3s) validate the reliability of our approach. 
% This suggests that superior occupancy reconstruction ensures rich feature representations, which in turn serve as a pivotal factor in boosting forecasting performance.
% Notably, our method outperforms OccLLaMA and delivers performance comparable to OccLLM. 
% These results imply that the proposed Cascade World Model is capable of effectively interpreting and forecasting complex driving scenes.
We evaluated our model on the Occ3D-nuScenes dataset for the forecasting task, predicting the future 3s conditioned on a 2s history. 
As reported in~\tabref{tab:forecast_planning_performance}, our method achieves significant gains over OccWorld, with improvements of 3.65\% in IoU and 3.2\% in mIoU. 
As shown in~\figref{fig:qualitative}, these quantitative gains translate into visually superior predictions that accurately capture the evolution of dynamic objects and fine-grained static details. 
Consistent improvements across all time steps (1s–3s) suggest that superior occupancy reconstruction ensures rich feature representations, which are pivotal for boosting forecasting performance. 
Notably, our method outperforms OccLLaMA and delivers performance comparable to OccLLM, demonstrating its capability in complex driving scenes.

\subsubsection{Motion Planning}
While precise forecasting of environmental dynamics is essential in autonomous driving, generating collision-free and reliable trajectories without manual intervention is of paramount importance. 
% To assess the reliability and robustness of our proposed method, we evaluated trajectory planning and collision rates on the nuScenes dataset.
Evaluated on the nuScenes dataset (\tabref{tab:forecast_planning_performance}), our method exhibits L2 errors comparable to prior arts with negligible margins. 
% However, recent studies such as BEVPlanner~\cite{li2024ego} have argued that L2 error metrics can be biased towards the benchmark distribution and may not fully reflect the true quality of planning. 
% Given that motion planning evaluation should prioritize safety, our achievement of state-of-the-art performance in collision rate is particularly noteworthy.
However, considering the potential bias in L2 error metrics noted by BEVPlanner~\cite{li2024ego}, our state-of-the-art collision rate is particularly noteworthy.
% Specifically, the significant performance improvements observed in the long-term horizon (2s–3s) demonstrate that our method ensures reliable planning even for distant future steps. 
% This result suggests that the high-fidelity reconstruction and accurate forecasting capabilities of our model effectively contribute to the downstream motion planning task.
Specifically, significant improvements in the long-term horizon (2s–3s) further demonstrate that our high-fidelity reconstruction and forecasting capabilities effectively ensure robust downstream planning.

\subsubsection{Ablation study of CascadeOcc}
% We conducted an ablation study on the nuScenes-Occ3D dataset to validate the efficacy of the proposed key components in CascadeOcc: the Cascade Occupancy World and TimeMixer. 
% As reported in \tabref{tab:ablation}, the Cascade Occupancy World improves mIoU and IoU by 2.36\% and 1.78\%, respectively, over the baseline. 
% This indicates that its intrinsic structural representation effectively preserves fine-grained details within complex driving environments. 
% Similarly, the TimeMixer achieves gains of 2.20\% and 2.23\% in mIoU and IoU. 
% This improvement stems from the effective aggregation of temporal hierarchies, which allows for the fusion of both long- and short-range historical contexts. 
% Finally, integrating both modules results in impressive improvements of 3.65\% in mIoU and 3.20\% in IoU. 
% This proves that the proposed components function synergistically to facilitate robust decision-making for autonomous driving.

We conducted an ablation study on the nuScenes-Occ3D dataset to validate the efficacy of CascadeOcc's key components: the Cascade Occupancy World(\secref{sec:cascade-tf}) and TimeMixer(\secref{sec:timemixer}). 
As reported in \tabref{tab:ablation}, the Cascade Occupancy World improves mIoU and IoU by 2.36\% and 1.78\%, respectively, indicating that its intrinsic structural representation effectively preserves fine-grained details. 
Similarly, TimeMixer achieves gains of 2.20\% and 2.23\%, attributed to the effective aggregation of temporal hierarchies that fuses long- and short-range contexts. 
Finally, integrating both modules yields impressive improvements of 3.65\% in mIoU and 3.20\% in IoU. 
This proves that the proposed components function synergistically to facilitate robust decision-making for autonomous driving.

\section{Conclusion}
\label{sec:conclusion}
In this letter, we rethink the fundamental representation capacity of Autoregressive Occupancy World Models, diverging from the recent trend of simply incorporating auxiliary knowledge such as Large Language Models (LLMs) or explicit action inputs. 
We propose CascadeOcc, a novel framework designed to maximize intrinsic representational power through a coarse-to-fine multi-scale VQVAE. 
Furthermore, by leveraging the proposed TimeMixer with a dual-hierarchy mechanism, our method achieves state-of-the-art performance in both occupancy forecasting and safety-critical motion planning. 
% For future work, we envision integrating recent advancements, such as LLMs, into CascadeOcc in a plug-and-play manner to further enrich semantic understanding. %원래 conclusion
% Finally, addressing the scarcity of open-source implementations, we will publicly release our full codebase to accelerate research in the field.
Although our method demonstrates robust performance, we note occasional object omission or flickering in highly dense environments. % rebuttal conclusion
For future work, we envision integrating recent advancements, such as LLMs, into CascadeOcc in a plug-and-play manner to compensate for these challenging cases and further enrich semantic understanding.

{
\balance
\bibliographystyle{IEEEtran}
\bibliography{egbib}
}

% \vspace{50pt}

\end{document}